\documentclass{article}

\usepackage{PRIMEarxiv}

\usepackage[utf8]{inputenc} 
\usepackage[T1]{fontenc}    
\usepackage{hyperref}       
\usepackage{url}            
\usepackage{booktabs}       
\usepackage{amsfonts}       
\usepackage{nicefrac}       
\usepackage{microtype}      
\usepackage{lipsum}
\usepackage{fancyhdr}       
\usepackage{graphicx}       
\graphicspath{{media/}}     
\usepackage{algorithm}
\usepackage{algorithmic}

\title{CAST: Character labeling in Animation using Self-supervision by Tracking
 }

\author{
  Oron Nir \\
  Reichman University , Israel \\
  Microsoft Corporation\\
  \And
  Gal Rapoport\\
  Reichman University , Israel\\
   \And
  Ariel Shamir \\
  Reichman University , Israel \\
}

\begin{document}
\maketitle

\begin{abstract}
    Cartoons and animation domain videos have very different characteristics compared to real-life images and videos. In addition, this domain carries a large variability in styles. Current computer vision and deep-learning solutions often fail on animated content because they were trained on natural images. In this paper we present a method to refine a semantic representation suitable for specific animated content. We first train a neural network on a large-scale set of animation videos and use the mapping to deep features as an embedding space. Next, we use self-supervision to refine the representation for any specific animation style by gathering many examples of animated characters in this style, using a multi-object tracking. These examples are used to define triplets for contrastive loss training.
    The refined semantic space allows better clustering of animated characters even when they have diverse manifestations. Using this space we can build dictionaries of characters in an animation videos, and define specialized classifiers for specific stylistic content (e.g.,\ characters in a specific animation series) with very little user effort. These classifiers are the basis for automatically labeling characters in animation videos. We present results on a collection of characters in a variety of animation styles. \newline Code and resources are available at: \url{https://github.com/oronnir/CAST}.
\end{abstract}

\keywords{Representation Learning \and Contrastive Learning \and Multi-Object Tracking \and Animation}

\section{Introduction}
\begin{figure}
    \centering
    \includegraphics[width = 7cm]{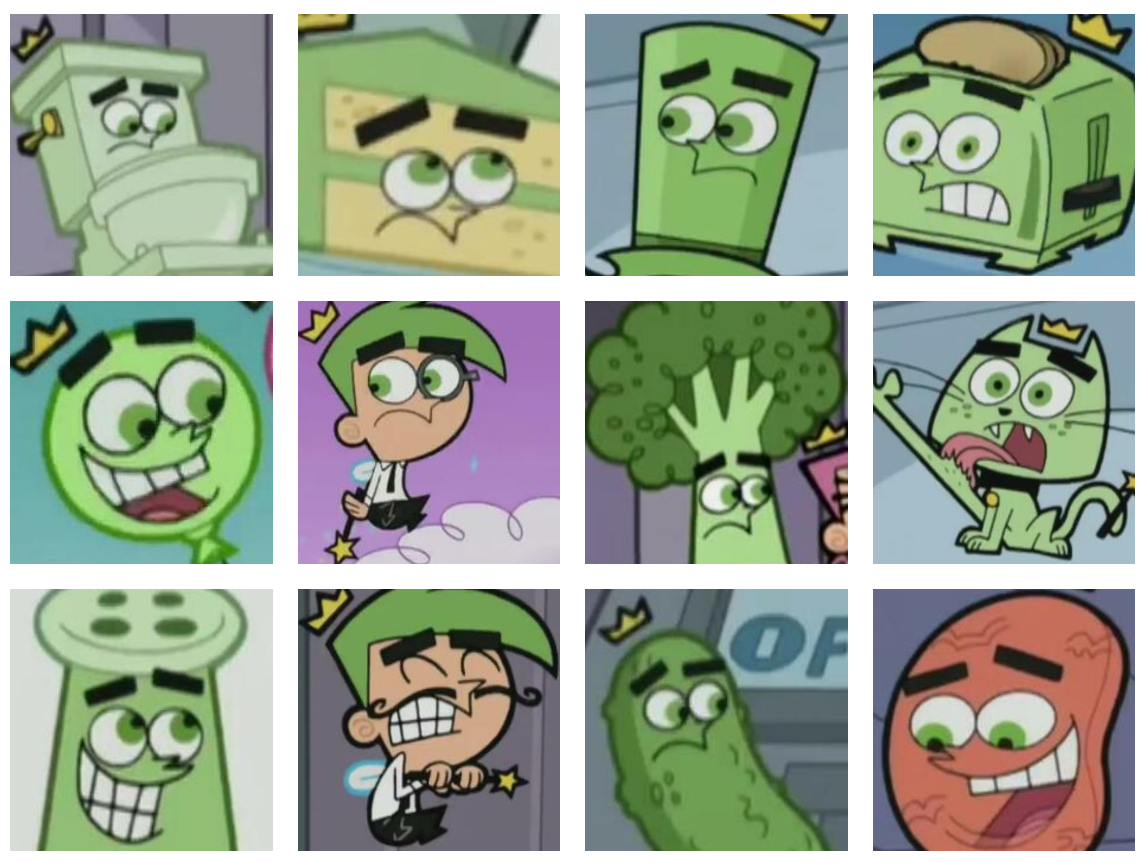}
\caption{The same character in animation can change color, appearance and even shape.
Our challenge is to build a representation that can maintain its semantics and map all visual depictions to the same identity.}
\label{fig:CosmoPhysicsViloation}
\end{figure}

Recognizing and labeling characters in a given video is an important task for many applications. It can empower media companies to search, analyze, and reuse content in videos. 
Many current solutions for real-life videos are based on face detection or people detection. However, in the animation and cartoon domain this task is more challenging. 
First, there is a wide diversity of styles and genres in animation such as Manga, hand-drawn and CGI movies that can have significant differences in appearance. In addition, they are all very different from real-life videos. Trying to build one semantic representation for all animation styles or relying on representations constructed for real-life videos is problematic.
Second, characters in animation videos could be anything from a person to an animal or even an object like a talking candelabra. Relying on face or people detectors is insufficient. Third, the same character can change color, texture, style, and shape, even in the very same scene (see Figure~\ref{fig:CosmoPhysicsViloation}).
Such variability means that the semantic representation of a character cannot be based on appearance alone, cannot use natural images learned representations and must include several possible instance variations of the same character.

In this paper we suggest learning a semantic representation suitable for character identification and labeling specific to a given animation style  using self supervision. As a first step towards generalization, we label an animated video dataset in various styles and leverage it to refine an object detector and to define a base latent representation for animation by refining a classifier in a supervised setup. This step is performed once to better capture the characteristics of animated media as opposed to natural videos for detection. On the other hand, character labeling demands precision for identification, and therefore a more style-specific representation.
To adapt the representation towards specific animation styles and move different instances of the same character closer in the semantic space we introduce a self-supervised learning method.
The method takes an input video (e.g., one episode of a series or a full feature film), segments its shots, and runs the base object detector on this video. 
A multi-object tracker (MOT) is later applied per shot to follow all characters in the shot through changes in appearance and shape.
Next, triplets of examples are sampled based on the characters' tracklets, and are leveraged to refine the learned representation using contrastive loss. The underlying hypothesis is that two detected characters within the same tracklet share identity while two that appears in the same frame do not. This alters the semantic space towards representing the specific animated style, where distances convey the identity of characters and not simply appearance differences.

To illustrate the effectiveness of the new representation space we use clustering to build more coherent and meaningful clusters of examples for each character. We demonstrate the application of this space for long-tail dictionary building of characters in a video, where many characters, even if they appear sparsely, populate the dictionary. We also use this representation to define more effective animated character classifiers, and demonstrate their use for dense identification and labeling of characters on unseen test-videos. 
Hence, we call our method CAST: Character labeling in Animation using Self-supervision by Tracking.

\section{Related Work}
\label{sec:related_work}

Recent advancements in deep neural networks applications for various computer vision tasks like object detection and facial recognition have matured to solve problems at an industrial scale. However, specific domains with less available data, like animation and art, many times require a different approach~\cite{ogawa2018object,yaniv2019face,tsubota2018adaptation}.

Animated character recognition has been addressed with machine learning models before. Yu et al.~\cite{yu2011fuzzy} have suggested a fuzzy diffusion distance as a cartoon similarity metric for recognition and clustering purposes achieving up-to 77\% accuracy, using Kuhn-Munkres algorithm, on a proprietary dataset with 60 classes. Zhang et al.~\cite{zhang20132} aimed at detecting IP violations and piracy with Scalable-Shape Context and Hough Voting. Nguyen et al.~\cite{nguyen2017comic} suggested a convolutional neural network-based (CNN) comic characters detector for a comic book recognition problem. They focus on a binary task of character detection and have reported 92\% precision. They have based their detector on the YOLO-V2 detector by Redmon \& Farhadi~\cite{redmon2017yolo9000}. Our work identifies characters in video, that introduces higher character variance compared to static comic books. Chu et al.~\cite{chu2017manga} have suggested a CNN based model for detecting Magna faces in grey scale. They have addressed this challenge and taken the approach of using a joint CNN for bounding box regression and non/character classification on candidate regions extracted using selective search. Their method reached 66\% F1 score on the \textsc{Magna109} dataset~\cite{fujimoto2016manga109}. Ogawa et al. ~\cite{ogawa2018object} have used an SSD300 architecture for the same detection problem on this dataset and reached an average precision of 67\% and 79\% on face and body, respectively.

Landmark detection in art was also addressed by ~\cite{yaniv2019face, jha2018bringing} and emotion recognition in animation by ~\cite{stark2018facial, yang2019human}. Caricature recognition was benchmarked by Huo et al. ~\cite{huo2017webcaricature} suggesting that the artistic aspects embedded in a caricature challenges the approach of augmenting real-life images as they impose a significant domain difference.

\begin{figure*}
\centering
 \includegraphics[width=\textwidth]{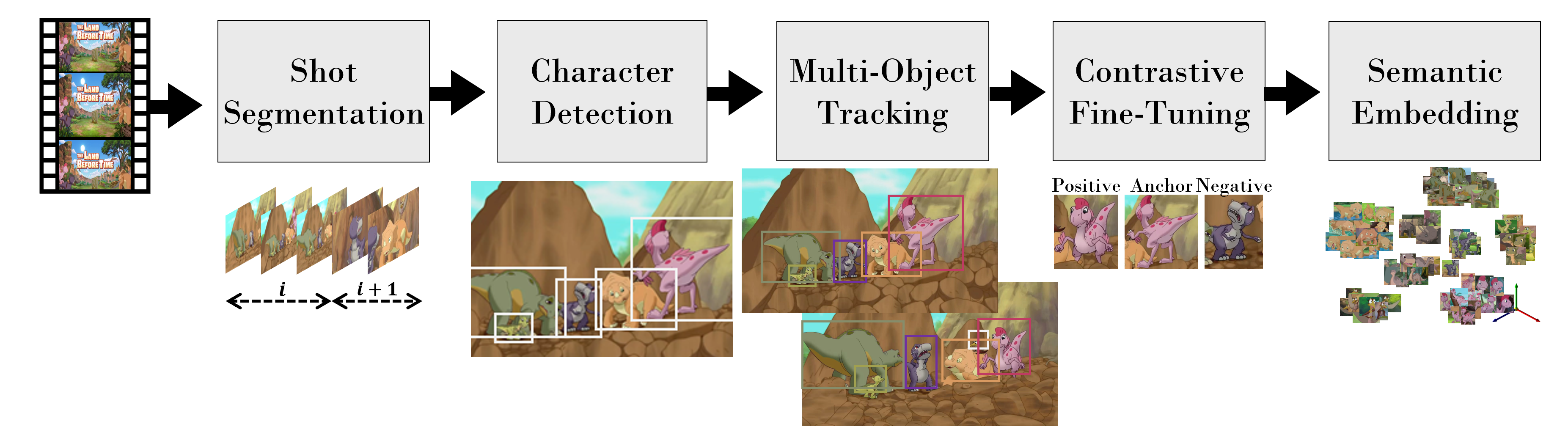}
 \caption{CAST self-supervised representation learning pipeline includes: automatic shot segmentation, detection, multi-object tracking, and contrastive-based refinement to define the unique semantic embedding for an animation style.}
\label{fig:pipeline}
\end{figure*}

Human character identification that works on realistic videos usually rely on detection and clustering of faces or humans~\cite{identification-09, Whosthat-16, azab-etal-2018-speaker, nameFace-19}. Naming of characters is usually done by searching the web or using social media. Since animation resources and datasets are less common, the only manual part in our method is naming characters for identification. 

Clustering of real-life faces is yet another well researched field. Zhu et al. ~\cite{zhu2011rank} have suggested a rank order distance to group similar instances of the same person. Schroff et al. ~\cite{schroff2015facenet} have introduced FaceNet and the triplet-loss for projecting images onto a latent space that quantifies similarity in a supervised-learning manner. Recently, Somandepalli et al.~\cite{somandepalli2021robust} used tracking of faces in a photo-realistic video, followed by clustering and verification using MvCorr\cite{somandepalli2019multiview} and Improved Triplet \cite{zhang2016deep} to adapt available face representation data to perform better on racially diverse images following \cite{sharma2019self}. Aneja et al. ~\cite{aneja2016modeling} have suggested DeepExpr model for facial expression recognition for multiple styles. Tsubota et al. ~\cite{tsubota2018adaptation} have used the \textsc{Manga109} dataset for manga face clustering. The approach they took for generalization of deep embeddings adaptation on the inferred manga comics was based on deep metric learning (DML) inspired by the work of Zhang et al. ~\cite{zhang2016joint}. They reached a normalized mutual information (NMI) of 71\% and accuracy of 64\% based on manga domain specific priors e.g., co-occurrence per comic scene. The `k' number of characters was assumed to be given and was not estimated. The underlying CNN architecture was ResNet50 ~\cite{he2016deep}. An ablation study suggested that without the page information the NMI drops to 67\% and the accuracy to 58\%. This indicates the importance of a temporal analysis in our challenge. Shen et al. ~\cite{shen2019discovery} have addressed a similar problem, discovering visual patterns in art collections with spatially consistent feature learning also known as ArtMiner. They have suggested a similarity metric based on ResNet18 for artwork elements to automatically recognize copied pieces of work created in different styles e.g., oil, pastel, water-color, etc. They have reported 88.5\% accuracy on LTLL~\cite{Fernando2015CVIU} and 85.7\% mAP on Oxford5k~\cite{philbin2007oxford}. 
Recently, \cite{zheng2020cartoon} have published iCartoonFace dataset for face detection and identification in animated images. While anthropomorphized faces tend to have the same facial features people have, an animated character's body shares less common attributes and perhaps makes the generalization a more challenging task.

The closest prior work to ours is \cite{8017484} who built a system that aims at finding the main cast of a given movie which is termed `unsupervised discovery of character dictionaries' . Our work not only builds the dictionary but allows to create classifiers that can recognize and index characters at frame-level and generalize to additional episodes of a given series. Their work applied the MultiBox detector on uniformly sampled frames followed by an object tracker, ImageNet FC7 layer for embeddings and Afinity Propagation for clustering. Their overall dictionary F1-Score is 72\% and 70\% cluster purity on a dataset of eight videos of \textsc{SAIL AMCD}. To evaluate the over/under clustering performance they've measured the median number of clusters per character which was 3. We consider this paper as our baseline for dictionary building and compare our results.
Kim et al.~\cite{kim2020character} have addressed the same task. Their main contribution was a character detector that adapts the animation style with a region-based Faster R-CNN following the work of \cite{wang2019towards}. The rest of their algorithm aligns with \cite{8017484}. However, since their \textsc{ABCD} data set is proprietary and they do not report cluster purity and median clusters per character it was difficult to compare to.

\section{Representation Learning}
\label{sec:embedding}

An overview of the CAST representation pipeline is represented in Figure~\ref{fig:pipeline}. First, the video is segmented to shots according to the method of Hua et al.~\cite{hua2004optimization}. Next, frames are sampled consistently on which our character detector is applied, providing character bounding box proposals. These proposals are then tracked while those who share a tracklet should appear closer in the embedding space than other proposals in the same shot for they belong to different characters. A set of triplets examples are then sampled and used to further refine the basic mapping network for the specific style of animation. This representation is later used for clustering, dictionary discovery and classifier definition of characters in this animation style.

\begin{table*}
\centering
\begin{tabular}{|l|cccccc|}
\hline
DATASET ROLE & STYLES[\#] & VIDEOS[\#] & KEYFRAMES[\#] & HOURS &	CHARACTERS[\#] & BOXES[\#] \\
\hline\hline
\textsc{Training}   & 61 &	174 &   118,751 &	65.6 &	549 &	257,706  \\
\textsc{Test}       & 50 &	 50 &    38,821 &	24.1 &	681 &	90,325  \\
\textsc{Evaluation} &  7 &	 14 &    17,583 &	11.0 &	49 &	50,300  \\
\hline
\end{tabular}
\caption {Statistics for the \textsc{Training}, \textsc{Test}, and \textsc{Evaluation} CAST datasets. \label{tab:datasets}}
\end{table*}

\subsection{Data Acquisition}

Building a CNN-based semantic representation must address both the scale and diversity of the data in the domain.
Thus, we gathered data sets of cartoon and animated content from various sources such as media companies and the web.
For the basic \textsc{Training} dataset, we collected 174 videos in 61 different styles including Anime, CGI, 2D cartoons, and more (see the Appendix). Examples of series collections in this dataset are: Blender (3D CGI), legacy Looney-toons videos published under public domain (2D hand drawn in color as well as gray-scale), Manga of various styles, Indian/Korean animated films, and more. From these videos more than 118K frames were extracted and annotated  manually. Characters in each frame were marked with a bounding-box and named (using Microsoft’s UHRS - like \cite{fu2019rekall}). This resulted in more than 250K ground-truth bounding-box instances of 549 distinct characters.

We used this basic \textsc{Training} dataset to define two neural networks: the first was trained to detect basic proposals for animated characters in animation videos while the second was trained as the basic mapping of proposals into a semantic embedding space, which is later refined using self-supervision for each specific animation style representation.

For testing the domain-adapted detector and selecting a clustering algorithm, we used a second \textsc{Test} data set. This set contained 50 videos with a total length of 24 hours and 681 characters. This dataset was collected from YouTube with various styles e.g. Gaming, amateur and professional productions content, and various other genres.

For evaluating the dense character identification application, presented later, we collected a third, \textsc{Evaluation}, dataset from seven different cartoon series, two episodes each.
One episode was used to guide the self-supervision and train classifiers, and the other was manually labeled and used for testing the classification results based on the first episode. This data set is 11 hours long and includes 49 characters. 
More details on the three sets can be found in Table~\ref{tab:datasets}. Note that the three datasets do not contain any overlap i.e., no character, episode or series appears in more than one dataset. We plan to publish a labeled dataset for future work comparison.

We also used the \textsc{SAIL AMCD} dataset \cite{8017484} to evaluate CAST for unsupervised discovery of character dictionaries and compare our results to~\cite{8017484}. This set contains eight evaluation videos that are full-length feature films of different animation styles. The videos were trans-coded to standard HD and sampled at 4 FPS.

\subsection{Basic Animated Character Detection}

Due to the high variance in characters' appearance, and the fact that animated characters are not necessarily humans, a simple person detector for character proposals will not suffice.
Our base detector is the YOLO V2 generic object-detector architecture ~\cite{redmon2017yolo9000} which was originally trained on ImageNet~\cite{deng2009imagenet} 
with millions of images. 
We use our \textsc{Training} set to fine-tune YOLO's Person class for 20 epochs and use it as a binary classifier, character vs. non-character. The person detector was chosen as many animation characters still tend to have a human-like attributes. Data augmentation methods were applied to avoid overfitting randomizing both vertical and horizontal flip as well as resize with factor range $[0.5, 2.0]$.
We also tried to fine-tune a different architecture based on Faster-RCNN~\cite{ren2015faster}, which is a two-pass CNN. However, we found our architecture, based on YOLO's fast low compute one-pass detector, to be faster and produced higher average precision.
We compare this detector to the following alternatives: using the YOLO V2 single class person detector, and using all combinations of the classes: Person, Animal, Mammal, and Toy as a detector. The detectors' effectiveness was evaluated in terms of Average Precision (AP) at a 50\% Intersection over Union (IoU) using the \textsc{Test} dataset. Our basic character detector has reached an AP of 61\% significantly outperforming the YOLO V2 person detector (AP=50\%) and the combined classes version (AP=50\%) in a wide range of animation styles.

\subsection{Base Embedding Network}

Feature maps in deep layers of networks such as VGG~\cite{Simonyan15} and ResNet18~\cite{he2016deep} have been shown to represent high-level semantic meaning in many previous works. However, such models were trained mostly on natural images, therefore could misinterpret animated content. 
Similar to the above approaches, to learn a representation suited for our domain, we use our \textsc{Training} dataset to fine-tune classification networks and use the feature map of its deepest hidden layer as the representation of the base embedding space.
The \textsc{Training} set containing 549 different characters was heavily long tailed:  40\% of the characters have less than 50 samples. Therefore, we have balanced the dataset during training by repeating rare characters’ images, such that per epoch, each character has at least 50 samples. To avoid overfitting, data augmentation transformations have been applied on the training data at random using the following methods: crop resize, Affine transform, color jitter, and horizontal flip.

We compared different configurations of three different architectures:
ResNet18~\cite{he2016deep}, SEResNeXt~\cite{hu2018squeeze} and ArtMiner~\cite{shen2019discovery}. SEResNeXt was found to perform best and therefore the last average pooling layer was used ($20^{th}$ squeeze and excitation block). Yielding vectors of size 2,048, as the basic semantic representation for animation characters bounding boxes. 

\begin{figure}
    \centering
    \includegraphics[width = 8.0cm]{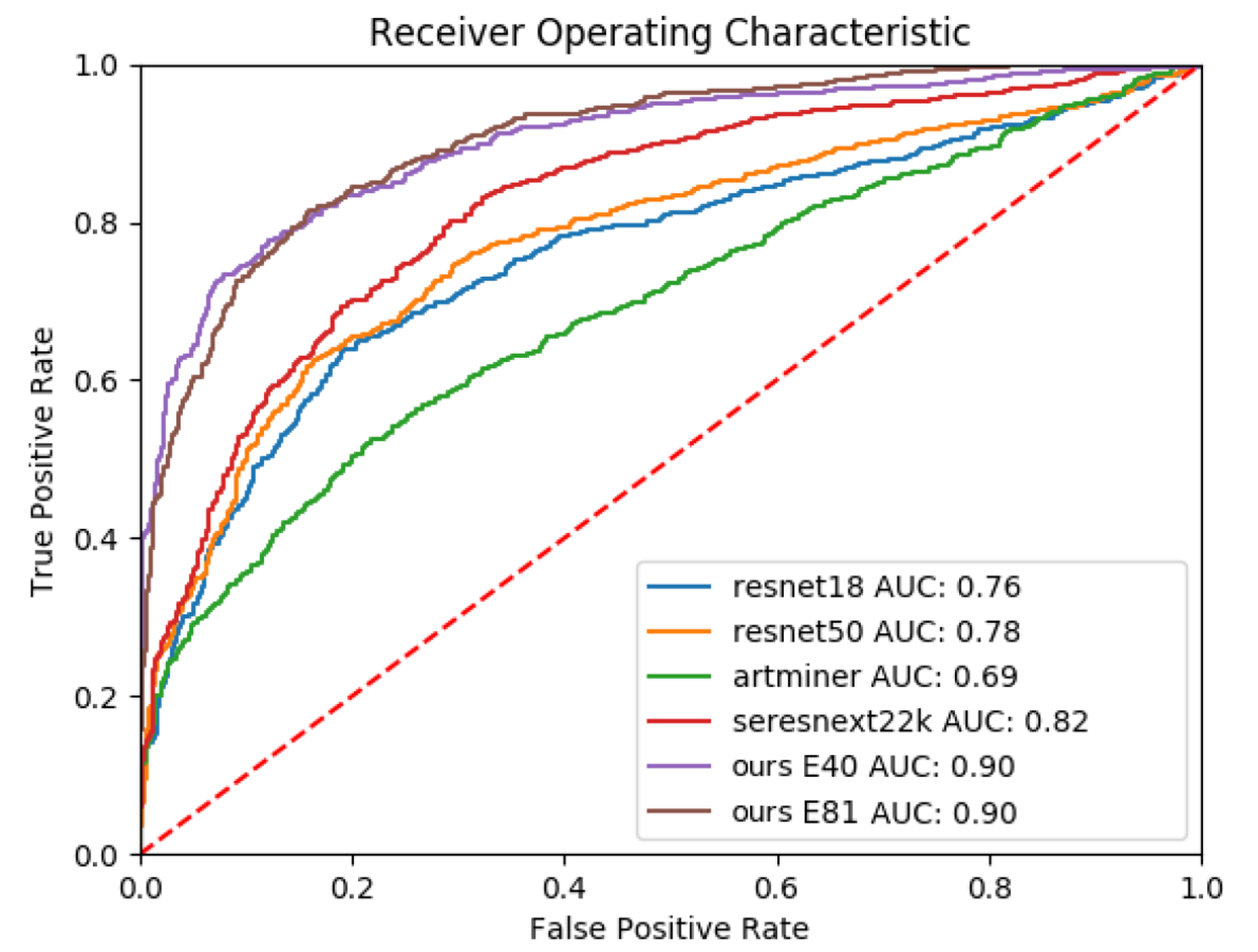}
\caption{RoC curve per basic semantic embedding backbone architecture: ResNet18, ResNet50, ArtMiner, SEResNeXt22k (the original network), our network fine-tuned for 40 epochs as well as a further trained version to illustrate the convergence. }
\label{fig:embeddings_RoC_artminer}
\end{figure}

To evaluate this representation, we measured how well this representation conveys semantic similarity. We built a test-set by creating a balanced dataset of 600k pairs of similar and dis-similar images of characters from the \textsc{Test} dataset. We defined a similarity measure for such pairs to be the cosine similarity of our embeddings of the extracted boxes. The pairs were ordered according to their similarity and addressed in a binary classification design with a RoC curve. In Figure~\ref{fig:embeddings_RoC_artminer} we plot the curves of the various tested alternative representations. Our representation, using an SEResNeXt back-bone, has outperformed all alternatives on the test data. We further found that training the network for more than 40 epochs does not provide much gain.

\subsection{Self-Supervision}

Similar to previous works, e.g.~\cite{kim2020character}, we strive to refine the semantic space for each specific animation style. Given a specific animation style (for instance, an episode of an animation series), we fine-tune the base embedding network towards this style, using a novel self-supervised approach which combines multi-object tracking with contrastive learning (triplet margin loss~\cite{schroff2015facenet, balntas2016learning} with margin=1.0). Our main hypothesis is that two samples from a single tracklet can be used as positive and anchor examples (coming from the same character but in different frames), while another character proposal from the anchor’s frame can be used as a negative example (coming from the same frame but different character).

\begin{figure}
    \centering
    \includegraphics[width = 8.0cm]{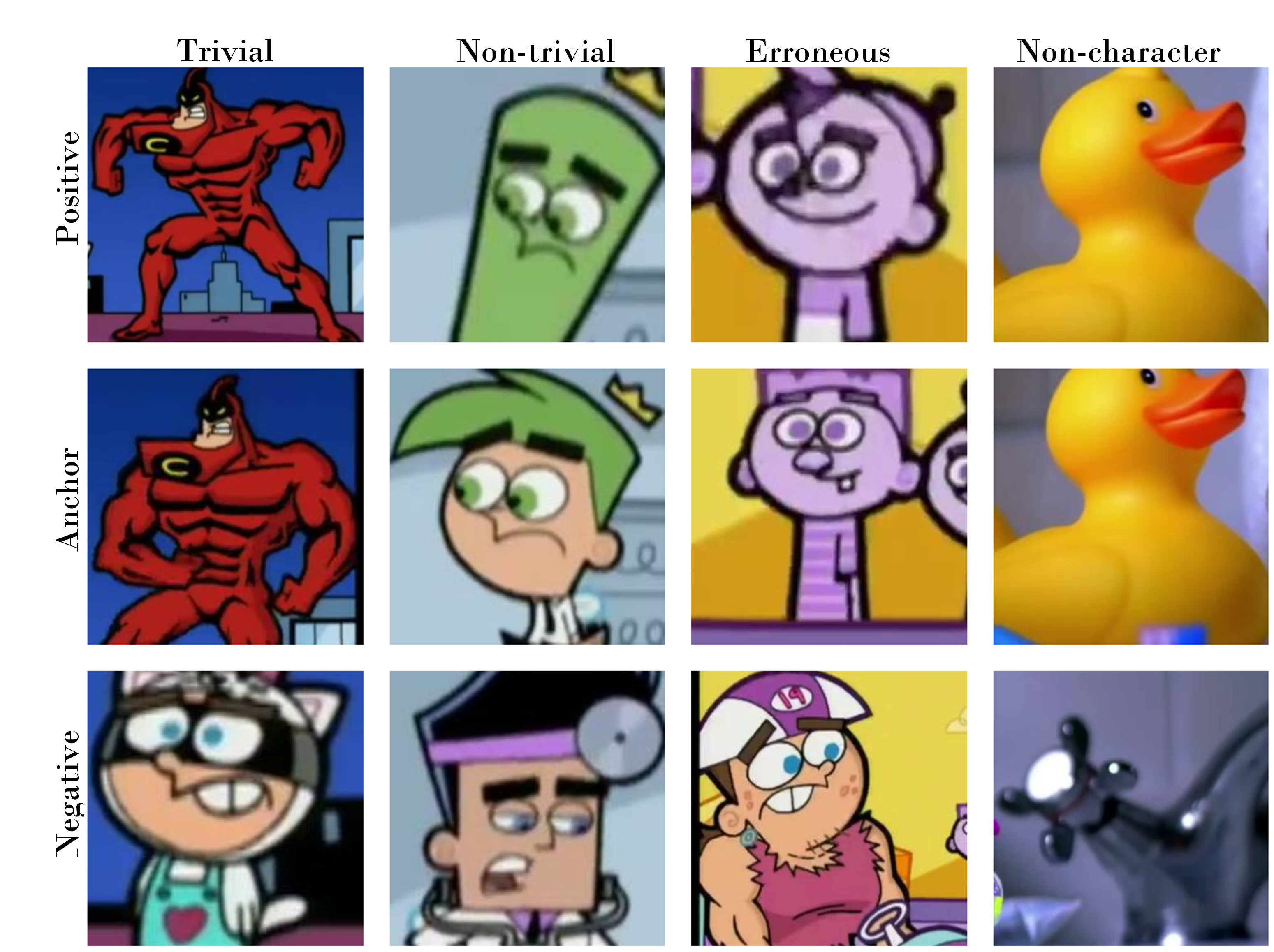}
\caption{Triplets examples illustrates the Positive, Anchor, and Negative triplets. The method aims for trivial and non-trivial examples while it is still resilient for erroneous and non-character examples.}
\label{triplets_examples}
\end{figure}

We sample 10k triplets and refine the base embedding network for 10 epochs using AdamW~\cite{loshchilov2018decoupled} ($LR=2\cdot10^{-5}$, batch size=20, $\lambda=10^{-4}$, $\gamma=0.1$, $\hat{m_t}=0.9$). This setup applies for any specific animation style in PyTorch~\cite{paszke2017automatic}. Tracking is performed with a classic multi-object tracking (MOT) approach of max-flow-min-weight algorithm inspired by the works of Zheng et al.~\cite{zhang2008global} and Wang et al.~\cite{wang2019mussp}. A sparse network based on detection is constructed, Next, an optimization algorithm finds the optimal flow in the graph that maximizes its MAP representation. Lastly, the tracks are derived from the flow function using a greedy shortest weighted path algorithm as described next.

To construct the network we include an artificial source and sink nodes, and each detected bounding box is considered as two nodes (`in' and `out') with a linking edge weighted by its detection confidence.  Skip connections are added every FPS frames to overcome temporary occlusions.
 The edge weight between detections of neighbor (or skip) frames is weighted by the following six factors: (1) Time gap proximity w.r.t. the sample FPS, (2) $P_{IoU}=0.5(IoU+1)$ so disjoint boxes could potentially be linked, (3) Scale difference ratio, (4) Euclidean distance in pixels of their centers, (5) semantic similarity using the original embeddings, and (6) scale-weighted center distance as characters which are closer to the camera may have higher relative angular velocity. The six factors above are aggregated together using a geometric weighted average into a match likelihood measure. Their corresponding weights are (1.0, 1.5, 1.5, 2.0, 3.5, 4.5).
Each edge in this network is considered as a unit capacity network and solved with the interior-point algorithm as a Linear Programming (LP) representation \cite{andersen2000mosek}. Since the constraints matrix is totally unimodular the LP solution is also a valid integer programming solution \cite{zhang2008global}.
Each tracklet solution per shot is greedily derived from the flow function using a weighted DAG shortest path algorithm and its MAP tracklet probability is considered to be its significance. Tracks significance filter threshold per shot is defined to be 10\% of the shot's most significant tracklet.

The triplets gathering is performed by sampling a shot with at least two tracklets, out of which a frame is picked, with at least two proposals (bounding boxes). The first is used as an anchor  while the second is used as the negative example. Then, from the anchor’s tracklet, a third proposal is randomly picked as the positive example. Figure~\ref{triplets_examples} illustrates the some triplets types. Many standard triplets are gathered using this method where a character preserves its appearance in the positive pair and contrasted with a different character. However, this method also captures non-trivial examples in which the character itself changes its appearance or shape. The two right columns in Figure~\ref{triplets_examples} illustrate erroneous triplets that can also be gathered using our method: the first is the results of errors in tracking causing a wrong positive pair, and the second is an error in the detector where non-characters (objects) are considered characters. However, due to the large amount of triplets gathered our mapping is resilience to such errors.

\section{Semantic Clustering}
\label{sec:clustering}

\begin{figure}
    \centering
    \includegraphics[width = 8.0cm]{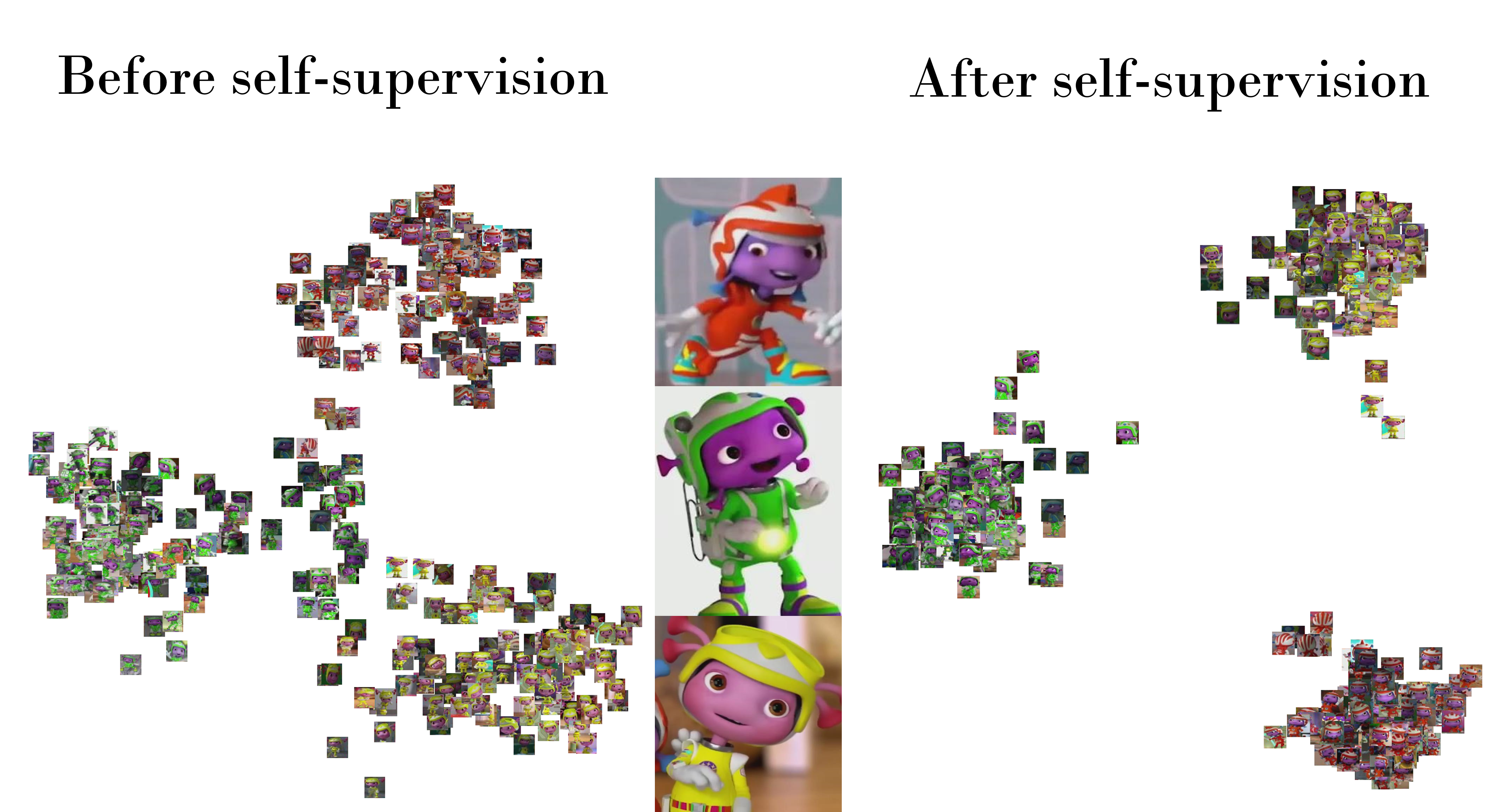}
\caption{Visualizing the first two principal components projection of our semantic embedding space before and after the Self-Supervision. We demonstrate the effectiveness of our method as a representation learning technique.}
\label{fig:clustering}
\end{figure}

A strong indicator for a meaningful representation is the ability to cluster items in the semantic embedding space.
Instances of the same animated character, even visually different, should all map to a relatively close position in the embedding space, while instances of different animated characters, even if they are visually similar, should be placed further away in the embedding space. Therefore, if we cluster the bounding boxes in the embedding space, the results should be coherent clusters of characters. 

First, we demonstrate the effectiveness of our base embedding network. We conducted an experiment, where we split the \textsc{Test} dataset to six subsets of various characters and used k-means clustering in the embedding space. 
We use cluster purity~\cite{manning2008introduction} as our measure as opposed to clustering evaluation measures like NMI and F-score, and evaluate the performance of the clustering quality as a function of the number of clusters.
We compared our embedding with SEResNeXt22, ResNet18, ResNet50, and ArtMiner for different values of `k'.
This comparison highlighted the superiority of our base embedding over the alternatives as for every value of `k', the purity of clusters in our base embedding was significantly higher. The full details of this experiment can be found in the Appendix.

Next, we compared several clustering algorithms, namely k-means, mean-shift, modularity maximization, affinity propagation, DBSCAN agglomerative clustering, and spectral clustering on our \textsc{Test} dataset. We use purity \cite{manning2008introduction} and K-metric, which is the geometric mean of the class purity and the cluster purity over the contingency matrix~\cite{pereira2011generic}. The two best performing were DBSCAN and agglomerative clustering. They yielded a median cluster purity of 96\% and K-metric of 40\%. We selected DBSCAN as the basic clustering algorithm as it handles noisy character proposals filtering inherently.
The hyper parameters that govern the confidence score for filtering were determined by grid search using the \textsc{Test} dataset to be \(\lambda_1=0.275, \lambda_2=0.4\).
To optimize clusters purity, we binary search for the $\varepsilon$ parameter of DBSCAN so that it maximizes the Silhouette times the non-noisy proposals in the range $25\leq k \leq 60$. This unsupervised criterion balances purity and cluster size.

Since we optimized for \emph{recall} during the proposal detection stage, there are still many outliers that are non-characters among the proposals. We remove outliers inside each cluster and remove low-confidence clusters altogether. This filtering is based on a confidence score that was calculated as a product of the detection's confidence and a squared exponential kernel of the sample’s distance to the cluster center: 
$$k_{\textrm{SE}}(x, \mu) = \exp\left(-\frac{(x - \mu)^2}{\lambda_1^2}\right)$$
where \(\mu\) is the cluster's median, \(x\) is the sample, and \(\lambda_1\) is a hyper parameter. A sample is considered an outlier when its confidence score is less than the cluster's \(Q_{25} - \lambda_2 \cdot IQR\) where \(Q_{25}\) is the $25^{th}$ percentile, \(IQR\) is the interquartile range and \(\lambda_2\) is another hyper parameter.
Finally, clusters with average centers that are up-to 0.7 cosine-similarity are merged back together.

\begin{table}
\centering
\footnotesize
\caption {Clustering ablation study on our \textsc{Evaluation} dataset. The better algorithm has less \textbf{Clusters} to label, higher \textbf{Pure} percentage, more \textbf{Characters per video}, and as many \textbf{Labeled boxes per video}.}
\begin{tabular}{l|ccccc} \hline
  Self- & Clusters & Pure & Characters & Labeled boxes\\
  Supervision & [\#]$\downarrow$ & [\%]$\uparrow$ & per video[\#]$\uparrow$ & per video[\#]$\uparrow$ \\ \hline\hline
    Before & 28.7 &	70          &    7.0          &	396.2\\
    After     & \textbf{20.0}  &	\textbf{90} &   \textbf{13.1} &	\textbf{599.3} \\ 
    \hline \end{tabular}
    \label{tab:clusterin_ablation}
    \end{table}

\begin{table*}[ht]
\centering
\begin{tabular}{|p{0.15\linewidth} | p{0.065\linewidth} p{0.065\linewidth} p{0.065\linewidth} p{0.065\linewidth} p{0.065\linewidth} p{0.065\linewidth} p{0.065\linewidth} | p{0.063\linewidth}|}

\hline
Self-Supervision& \#1  & \#2    & \#3    & \#4    & \#5    & \#6    & \#7    & Avg. \\
\hline\hline
Before & 0.156 & 0.258 & 0.166 & 0.285 & 0.231 & 0.382 & 0.247 & 0.246\\
After  & 0.320 & 0.423 & 0.364 & 0.601 & 0.248 & 0.520 & 0.332 & 0.401\\
\hline
Silhouette gain$\uparrow$   & 0.164 & 0.166 & 0.199 & 0.317 & 0.017 & 0.137 & 0.085 & 0.155\\
\hline

\end{tabular}
\caption {Self-Supervision introduces a significant gain of 0.155 ($P$-val=0.0047) to the Silhouette score on our \textsc{Evaluation} dataset.}
\label{tab:silhouette_clustering_ablation}
\end{table*}

Our combined MOT and contrastive-loss based self-supervision allowed further representation enhancement.
The performance analysis using the \textsc{Evaluation} dataset shows that 90\% of clusters created after the fine-tune stage were totally pure across all styles, while each character has a median of 1.2 clusters (lower is better). This demonstrates a significant improvement over the base-embedding as the clustering algorithm has yielded 30\% less clusters for the same video i.e., lower over segmentation. These clusters captures 13.1 characters per video instead of 7.0 i.e., 87\% better while these clusters have higher purity.  (See Table~\ref{tab:clusterin_ablation} and an illustrative example in Figure~\ref{fig:clustering}). 
As can be seen in Table~\ref{tab:silhouette_clustering_ablation}, for all different style of videos in our \textsc{Evaluation} set, the clustering quality measured using silhouette increases as a result of our self-supervision refinement.

\section{Applications}
\label{sec:applications}

\subsection{Long-Tail Dictionary Construction}

The first step towards automatic labeling of characters in animation movies is automatic characters cast discovery. In this task, the goal is to build a dictionary of characters appearing in the movie. For any given animation style, we first use our automatic method to learn the refined mapping tuned to this style. Then, for each new episode or movie in this style we use our detection pipeline to gather character proposals and map them to the semantic space. We then cluster and filter the proposals in this representation space.
This representation promotes purity so that each cluster will contain proposals belonging to a single character and encourages only a small number of clusters (preferably one) representing the same character. Then, to build the dictionary we create an entry for each significant cluster.

\textbf{Proposals Creation:} To reduce the chance of a miss-detection, the character proposal detector is tuned to be \textit{recall} oriented as False-Positive detection is recoverable downstream while False-Negative is not. Only bounding boxes with less than 20\% confidence or having a size smaller than 2.5\% of the frame are filtered out at this stage.
This causes repetitions of semi-identical proposals, i.e., bounding boxes with highly similar content. Hence, we apply a screening procedure to remove duplicate proposal. Each proposal is represented by an Edge Directional Histogram (EDH) feature vector of size 116~\cite{wang2012duplicate}. The 64-dimensional color and texture features ensures global similarity between two proposal and the \(4 \cdot 13\) EDH features ensures spatial similarity with detailed constrains by edge. The EDH was computed using a Canny edge detector with a \(7\times7\) Gaussian convolution kernel.
Next, the cosine-similarity was computed between all pairwise proposals, and an undirected graph was built with proposals as nodes and cosine-similarity as edge weight. Edges of weight lower than \(0.995\) were pruned. Finally, cliques were found in this graph using~\cite{bron1973finding, CAZALS2008564}, and each clique was aggregated into a single proposal.

\textbf{Characters Selection:} 
To build the dictionary we first cluster all candidate bounding boxes and filter them as described in Section~\ref{sec:clustering}. Next, we select every cluster and compute the median vector of all proposals in the cluster. We pick the proposal closest to the median as the representative and insert it to the dictionary.
We report the performance of our algorithm for unsupervised discovery of character dictionaries both on our \textsc{Evaluation} dataset and on \textsc{SAIL-AMCD}, comparing our results to~\cite{8017484}. Although \cite{kim2020character} reported similar results to ours, their dataset was not available and they did not report the aggregated purity of their clusters.

\begin{table*}[ht]
\centering

\begin{tabular}{|p{0.18\linewidth} |p{0.07\linewidth} p{0.05\linewidth} p{0.05\linewidth} p{0.06\linewidth} p{0.12\linewidth} p{0.12\linewidth}  p{0.11\linewidth}|}\hline

SAIL AMCD Video /Method avg. score & Precision\newline [\%]$\uparrow$ & Recall\newline[\%]$\uparrow$ & F1 \newline[\%]$\uparrow$ & Purity \newline[\%]$\uparrow$ &	Med.\newline exemplars per character$\downarrow$ & Avg.\newline exemplars per character$\downarrow$ & Additional \newline characters [\#]$\uparrow$\\
\hline\hline
Cars 2	                    & 76.9	& 75.0	& 75.9	& 98.1  & 1.00	& 1.286	& 13\\
Free Birds	                & 100.0	& 90.0	& 94.7	& 98.6  & 1.00	& 1.360	& 15\\
Frozen	                    & 96.0	& 100.0	& 98.0	& 99.5  & 2.00	& 2.000	&  9\\
Dragon 2                	& 95.0	& 91.7	& 93.3	& 94.7	& 1.00	& 1.444	&  6\\
Shrek Forever After         & 89.7	& 90.0	& 89.8	& 90.5	& 1.00	& 1.679	& 18\\
Tangled	                    & 86.4	& 77.8	& 81.8	& 84.2	& 1.00	& 1.714	&  5\\
The Lego Movie              & 85.2	& 100.0	& 92.0	& 95.0	& 1.00	& 1.522	& 11\\
Toy Story 3	                & 93.5	& 77.8	& 84.9	& 99.8	& 1.00	& 1.483	& 11\\
\hline
CAST (ours)                 & \textbf{90.3} & \textbf{87.8} & \textbf{88.8} & \textbf{95.1} & \textbf{1.125} & \textbf{1.561} & \textbf{11.0}\\
\hline\hline
Somandepalli et al.   & 81.0  & 65.2  & 72.2  & 70.3      & 3.00  & N/A   & N/A\\
 \hline

\hline
\end{tabular}
\caption {Unsupervised Discovery of Character Dictionaries comparison on the \textsc{SAIL AMCD} test videos.  
\label{tab:SAIL_evaluation}}
\end{table*}

On our CAST \textsc{Evaluation} dataset, yields 16.8 clusters per video, when the average number of characters per video is 13. On average the cluster purity was 98.5\%. 
The precision, recall and F1-score of these dictionaries are 91.8\%, 83.1\%, and 86.6\%.
A visual illustration of the power of self-supervision refinement can be seen in Figure~\ref{fig:CosmoPhysicsViloation}. Not only that all the different manifestation of this character (including kid/toaster/cat/broccoli) were mapped to the same cluster, but this cluster, including 94 bounding boxes, was 100\% pure.

Comparing CAST with the work of Somandepalli et al.~\cite{8017484} on the \textsc{SAIL AMCD} test set (see Table~\ref{tab:SAIL_evaluation}). CAST outperforms the state of the art in all metrics and indicates 16.6\% improvement in F1 as well as 24.8\% in purity. Moreover, CAST allows to build a much larger dictionary containing twice as many characters that were not chosen in the original evaluation as lead characters (see examples of dictionaries in the Appendix).

\subsection{Defining Classifiers for Characters}

The second step towards automatic labeling of characters in animation movies is the construction of dedicated classifiers for each character in the movies. The key idea is to gather enough examples that are diverse enough for each character to build a training set for effective learning. Using CAST, the dictionary is presented to the user for naming. This allows us to use all the proposals in every named cluster as training data for classifiers. This also allows us to merge clusters when the same character is found more than once in the dictionary. Lastly, this also allows us to filter noisy clusters, where several characters appear in the cluster. We do this by presenting a number of examples from each cluster to the user and validating that they are all examples of the same character.

Using this method, a specialized training set for each character can be constructed with minimal user effort. These training sets are then used to train specialized classifiers as needed. 
Moreover, when processing a new video, the user can choose to apply the existing classifier model for automatic identification of some characters or add additional new characters by further naming clusters of unidentified characters.

\textbf{Training Classifiers:}
All the named clusters are merged into one training-set that allows training a multi-class image classifier for the specific animation styles. In our experiments we fine-tune a state-of-the-art CNN classifier SEResNeXt~\cite{hu2018squeeze} for 40 epochs using typical augmentation operations such as rotations and mirroring. 

To best train a mutli-class classifier, it is important to introduce an additional class of negative examples which comes from the same animation style. Typically, these examples are cropped from the background of processed images. Since our model is aimed at working in the wild on new series and new animation styles, we devised a method to create negative examples automatically per animation style. 
The base detector that marks characters bounding boxes on keyframes is biased for recall. Hence, our hypothesis is that all other pixels can be considered background. The challenge is to find large enough sub-frames that do not intersect with the characters' bounding boxes (examples available in the Appendix). Finding the largest empty rectangle (LER) is a known problem \cite{atallah1986note}. In practice, we devised a recursive algorithm that extracts more than a single empty rectangle per keyframe, not necessarily the largest ones, but with some minimal width, height, and area at a runtime of $O(n^2)$ where $n$ is the number of proposals. For the algorithm and the proof see the Appendix.
The resulting classifier can then be used on other videos of the same style (e.g.,\ other episodes of the same animation series) to classify known characters. Still, new characters can appear in such videos, as well as some known characters with different appearances. In such cases, our method can be applied again with the same procedure of clustering and naming but only on the unknown proposals of any new video processed, to create new training examples and re-train the classifier further.

\begin{table*}

\centering
\begin{tabular}{|p{0.22\linewidth} | p{0.07\linewidth} p{0.07\linewidth} p{0.09\linewidth} p{0.1\linewidth} p{0.1\linewidth} p{0.07\linewidth} p{0.07\linewidth}|}
\hline
  	                            & Detected	  &	 Clusters & Number	of  & Clusters  per & Boxes per &Relevant & General \\
Series (credit \textcopyright)	&	proposals &	 to name  & characters  & character	    & character &Purity[\%]&Purity[\%]\\
\hline\hline
1. Bob the builder (HIT)	        &	 2,648    &	20	      &   10        & 1.7           & 26.3 &  95.5 & 95.5\\
2. Fairly odd parents\newline (Nickelodeon) &	 4,215    &	17	      &	  18        & 1.0	        & 44.3 &  99.1 & 99.1\\
3. Fireman Sam (HIT)	            &	 4,633    &	14        &	  14        & 1.5       	& 35.4 &  97.2 & 79.1\\
4. Floogals (Jellyfish Pic.)	    &	 4,163    &	4	      &	   4        & 1.0	        & 92.8 &  99.2 & 99.2\\
5. Garfield (Mediatoon)	            &	 4,959    &	13	      &	   4        & 1.4	        & 53.5 &  99.5 & 99.5\\
6. Southpark (Viacom)	            &	 5,639    &	25	      &	  21        & 1.0	        & 16.6 &  100.0 & 93.8\\
7. Land before time\newline (NBCU)          &    4,795    &	11	      &	   8        & 1.5	        & 54.5 &  98.8 &98.8\\
\hline
\end{tabular}
\caption {Training videos statistics of seven different animation styles in our \textsc{Evaluation} set that were used for building the classifiers. CAST brings down user effort bounding boxes naming by two orders of magnitude, while still creating a sufficient training-set.  \label{tab:pipeline}}
\end{table*}

\begin{table*}[ht]
\centering

\begin{tabular}{|p{0.175\linewidth} | p{0.12\linewidth} p{0.12\linewidth} p{0.09\linewidth} p{0.12\linewidth} p{0.09\linewidth} p{0.1\linewidth}|}
\hline

Series name & Characters [\#] & Precision [\%] & Recall [\%] &	Accuracy [\%] & F1 [\%] & Support [\#] \\ 
\hline\hline
1. Bob the builder	    &	6	&	97.4	&	97.4	&	97.4	&	97.4	&	230 \\
2. Fairly odd parents	&	6	&	93.8	&	92.3	&	92.3	&	91.9	&	246 \\
3. Fireman Sam	        &	14	&	73.1	&	66.6	&	70.1	&	66.7	&	490 \\
4. Floogals	            &	4	&	94.7	&	92.6	&	93.9	&	92.6	&	163 \\
5. Garfield	            &	4	&	93.7	&	87.3	&	87.3	&	88.0	&	165 \\
6. Southpark	        &	7	&	94.5	&	94.1	&	94.1	&	94.2	&	104 \\
7. Land before time	    &	8	&	92.2	&	91.3	&	91.3	&	91.3	&	344 \\ \hline
Total	                &	49	&	91.3	&	88.8	&	89.5	&	88.9	&	1742\\
\hline
\end{tabular}
\caption {Results on test videos per style of our \textsc{Evaluation} dataset.  
\label{tab:evaluation}}
\end{table*}

\begin{figure}
    \centering
        \includegraphics[width = 8.0cm]{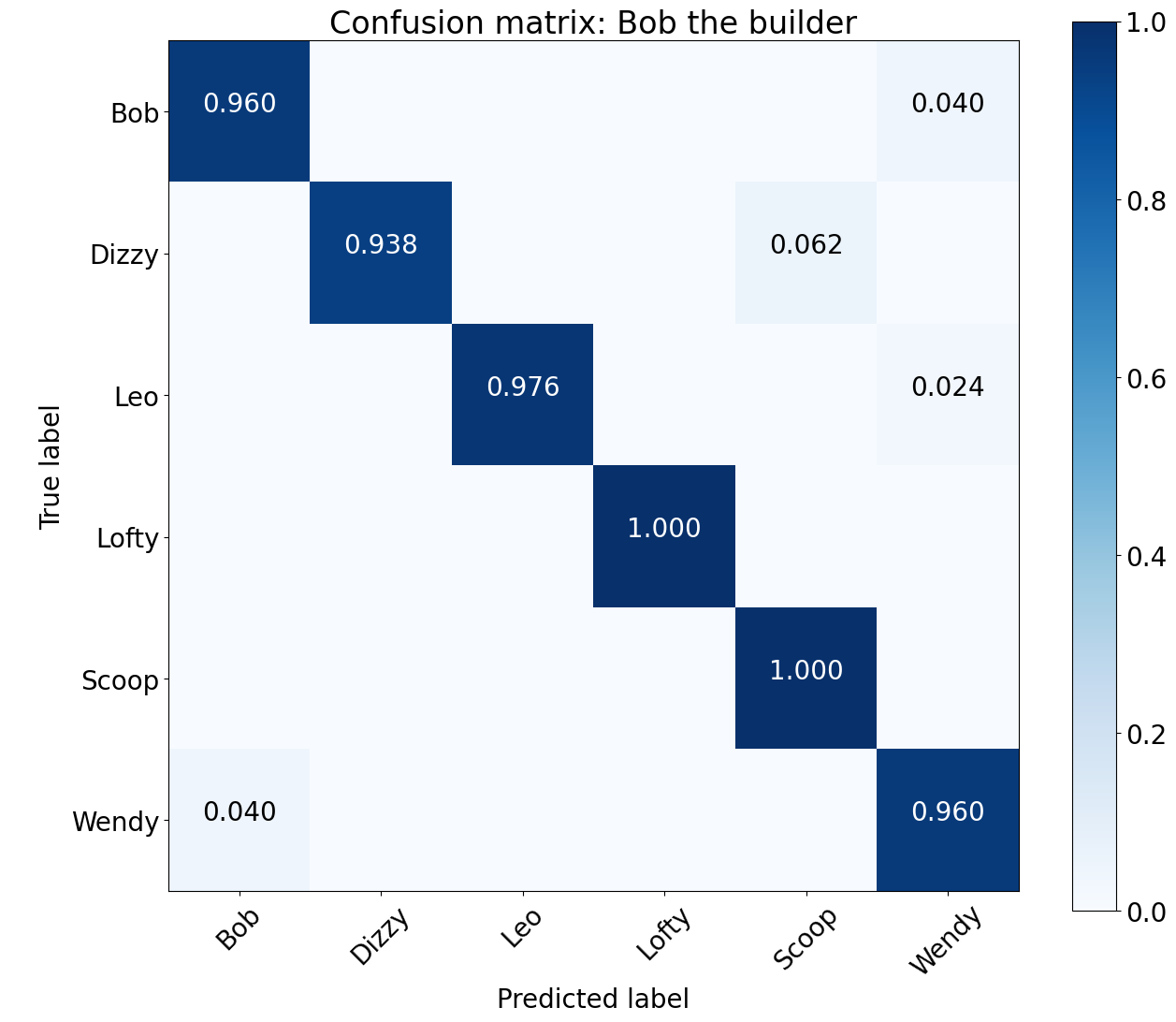}
\caption{`Bob the builder' confusion matrix (normalized rows).}
\label{SouthparkConfusionMatrix}
\end{figure}

\textbf{Experiments:}
To evaluate the effectiveness of CAST to define  specialized classifiers for animated characters identification, we use the \textsc{Evaluation} set that combines two episodes of seven different animation series. The animation styles include CGI, 2D, and Cutout techniques, with characters including humans, animals, dinosaurs, cars and tractors. For each series, we denote one training episode to create the dataset and train a classifier using CAST, and one test-episode, where several character proposals were manually selected and named for evaluation.

Statistics on CAST results for the seven train-episodes can be found in Table~\ref{tab:pipeline} and the Appendix. As can be seen, CAST allows defining classifiers for animation characters by naming (or discarding) just a few clusters from the dictionary instead of manual labeling and filtering thousands of proposal bounding boxes. Cluster Purity is reported both before and after the noisy cluster filtering under `General Purity' and `Relevant Purity' respectively. `Fireman Sam', staring four firemen figures which wear the same uniform sometimes clustered together and damage the general purity.

The results on the test-episode of each animation style can be found in Table~\ref{tab:evaluation}. 
There are differences in the number of appearances of the different characters in the seven test-episodes. We randomly picked, on average, 22 instances of each character in the test-episodes, creating 1,742 characters of ground-truth test data.
The total results yielded a mAP, F1-Score, and Accuracy of 92\%, 88\%, 89\% respectively. The best performing image classifier was created for the series `Bob the builder' with Accuracy and F1-score of 97.4\% and 97.4\% (see the confusion matrix in Figure~\ref{SouthparkConfusionMatrix}, and the  matrices of all series in the Appendix). The least performing image classifier was created for the series `The Land Before Time', still achieving mAP and F1-score of 82.4\% and 74.2\% on 8 characters.

We further tested two series by annotating \emph{all} proposal bounding boxes in the test-episodes of the series as ground truth. In this case, many proposals do not contain one of the characters and are marked as `unknown'. In the Appendix we show the confusion matrices in these cases. The matrices show almost no confusion among characters, but there are some mis-classification between characters and non-characters. There are fewer false-positive examples where non-character proposals are classified as characters and more false-negative examples. Error analysis on the false-negative cases revealed that many of them contain character parts such as hands, legs etc. These proposals were identified and annotated as characters by the human annotator but are still a challenge for automatic classifiers. 
This may be an avenue of future research on identifying partial and occluded animated characters. 

We confirmed an underlying assumption that training a classifier on different animations, consist of different characteristics, such as style, texture, colors, geometry as well as coarser and fine-grained characteristics, would yield different classifiers. By allowing a self-supervised training per animation, we demonstrated an overall improvement. Introducing a novel animation specific classifier framework, has significantly improved the classifier's  quality metrics (Purity, F1-Score, etc.). Moreover, it resulted a substantial reduction of $56.6\%$ in the number of clusters per character. That is, significant reduction in the annotations process.

\textbf{Class-per-cluster vs. Class-per-character:}
The method we used above aggregates all clusters that share the same label into a unified class for training. An alternative design can create a class per cluster to train the multi-class classifier and consolidates the class name results after the prediction. For instance, if the character Bob from the series `Bob the builder' has two dictionary entries, then their clusters will be used separately in the classifier training as Bob-1 and Bob-2, but both will be mapped to Bob after the prediction.
The hypothesis behind this alternative design is that the feature space might not be connected in terms of the label distribution in the high dimensional space. Consolidating all labeled data into the same class may introduce an unnecessary error. Alternatively, the clusters themselves encode connected (or convex) similarity regions that can represent the character in some particular settings like wearing  specific clothes.
To test this hypothesis a two-tail paired t-Test was conducted to compare the F1-score change in the class-per-character baseline design vs. the class-per-cluster alternative hypothesis.
The unit of analysis was a character using the \textsc{Evaluation} dataset (N=49). We found that the our choice of merging clusters performed 5.5\% better (higher) than the alternative (P-value=3.4\%). To conclude, our learned embedding-model seemed to encode coherent regions in a way that a consolidated class design would leverage better than separated classes per cluster.

\subsection{Dense Character Labeling and Statistical Analysis}

Once classifiers are defined for a specific animation style (e.g., series), any new video with the same style can be densely annotated using the classifiers on every frame. Some results showing these examples are shown in the supplemental video. Note that these were created without temporal coherency for tracking - only running the classifiers on every frame and automatically labeling each character classified. 

Using CAST, creative studios and media companies can benefit from a service that can automatically index and expose information both at a video and series level. Labeling each frame in the video with the characters that appear allows to gather important statistics on episodes as well as whole series.
This can assist animation productions in data management especially as more and more animation content is being created. For example, gender bias has become an important topic which affects us daily. By knowing the characters' gender and running a dense identification we can tell what is the screen time of each character and indicate, for instance, that the `Cars 2' movie introduces a gender bias of 78\% Male screen time.

\section{Discussion}

We have presented a method to learn a style-specific semantic representation more suitable for animated content using self-supervision. The self-supervision is based on multi-object tracking and building a dataset of triplets to refine a base mapping to the specific style. 

Using clustering in the semantic representation space allows us to automatically build character dictionaries for animation movies, to gather training sets for multi-class classifiers of characters, and eventually to automatically perform dense labeling of characters in animation videos. 
Such a solution allows users and media companies to analyze, reuse, search, and monetize animation content much more easily. 

\textbf{Limitations and Suggestions for Further Research:}
One limitation of CAST is that it relies on the initial detector for proposals. Although we tuned the detector for \emph{recall}, if the detector fails to mark the bounding box of a character as a proposal, the following steps cannot correct this.

\begin{figure}
    \centering
    \includegraphics[width = 8.0cm]{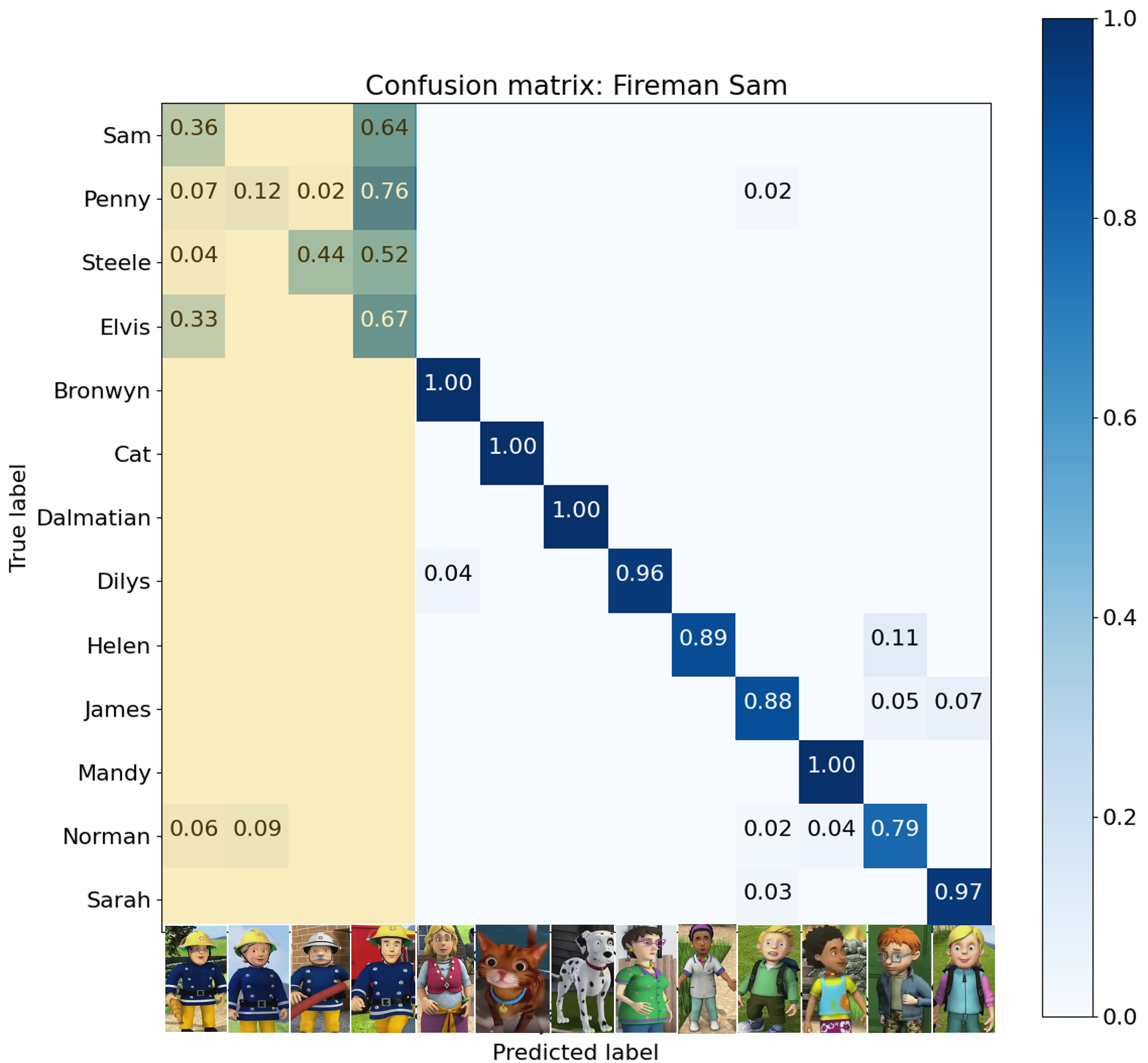}
\caption{Limitation in clustering characters that wear uniform. The four firemen cause over 80\% of the error.}
\label{FiremanSamLimitations}
\end{figure}

An underlying assumption of this research is that different animated characters proposals could be distinguished in the embedding space. 
However, since the network embedding still depends on visual features, 
when a subset of characters share common visual features, the embedding may still confuse between them. As a result, the classifiers can also be confused. An example can be seen in `Fireman Sam' series videos. 14 characters were recognized in this series, while only four of them are responsible for over 80\% of the classification error. These characters are the firefighters, and the obvious reason for confusion is their similar uniform (see the confusion matrix in Figure \ref{FiremanSamLimitations}). 

Our final character labeling did not take advantage of the temporal coherency of the video, but rather treated each frame separately. A more elaborate solution can be defined using smoothing and inference along with tracking.

Style variation in animated content is still extremely large. We refine our basic representation towards specific animation styles using embedding. Another possibility would be to combine such representation with domain adaptation techniques for specialized types of animation (see e.g.,\ Tsubota et al. \cite{tsubota2018adaptation} on manga comics).

In terms of building the classifiers, our evaluation tested only the two alternatives of class-per-character and class-per-cluster for building the training set. There is a whole range of possibilities between these two extremes that can combine clusters based on some cluster similarity measures to create the training set for characters.

\section*{Acknowledgements}
\label{sec:acknowledgements}
This research was partly supported by the Israel Science Foundation (grant No. 1390/19) and The Ministry of Innovation, Science and Technology (grant number 16470-3).
The authors would like to thank Maria Zontak, Apar Singhal, Lei Zhang, and Ohad Jassin (Microsoft) for their support of this research \cite{US20210056313A1,US20210056362A1,US10560734B2,US10902288B2}.

\bibliographystyle{unsrt}  
\bibliography{references}

\section*{Appendix}
\begin{figure*}
    \centering
    \includegraphics[width=\textwidth]{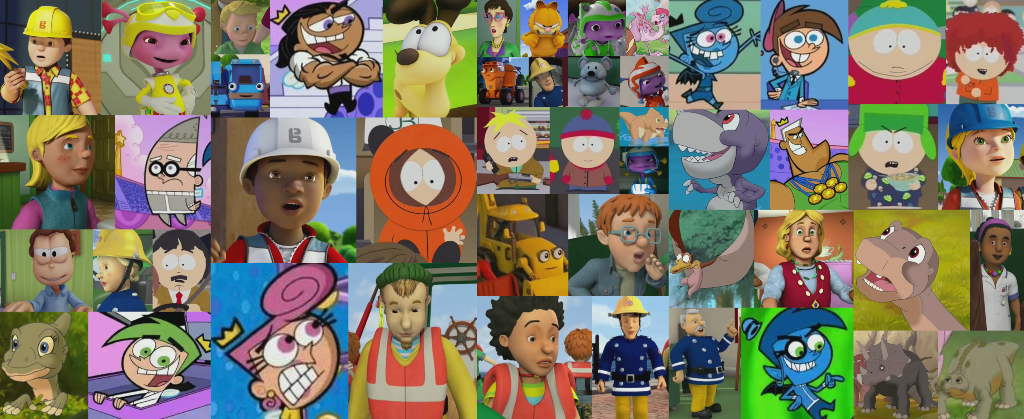}
\caption{All characters of our \textsc{CAST Evaluation} data-set from the series `Bob the Builder', `Fairly Odd Parents', 'Fireman Sam', `Floogals', `Garfield', `Southpark', and `The Land Before Time'.}
\label{All_characters_Evaluation}
\end{figure*}

Visualizing the challenge of generalization across animation styles using semantic learning representations on Figure\ref{All_characters_Evaluation} showing characters from our \textsc{CAST Evaluation} dataset with both dinosaurs, aliens, and tractors produced in CG, Cutout, and other animation technologies.

\section{Proposal Clustering}
\begin{figure*}
    \centering
    \includegraphics[width = 17cm]{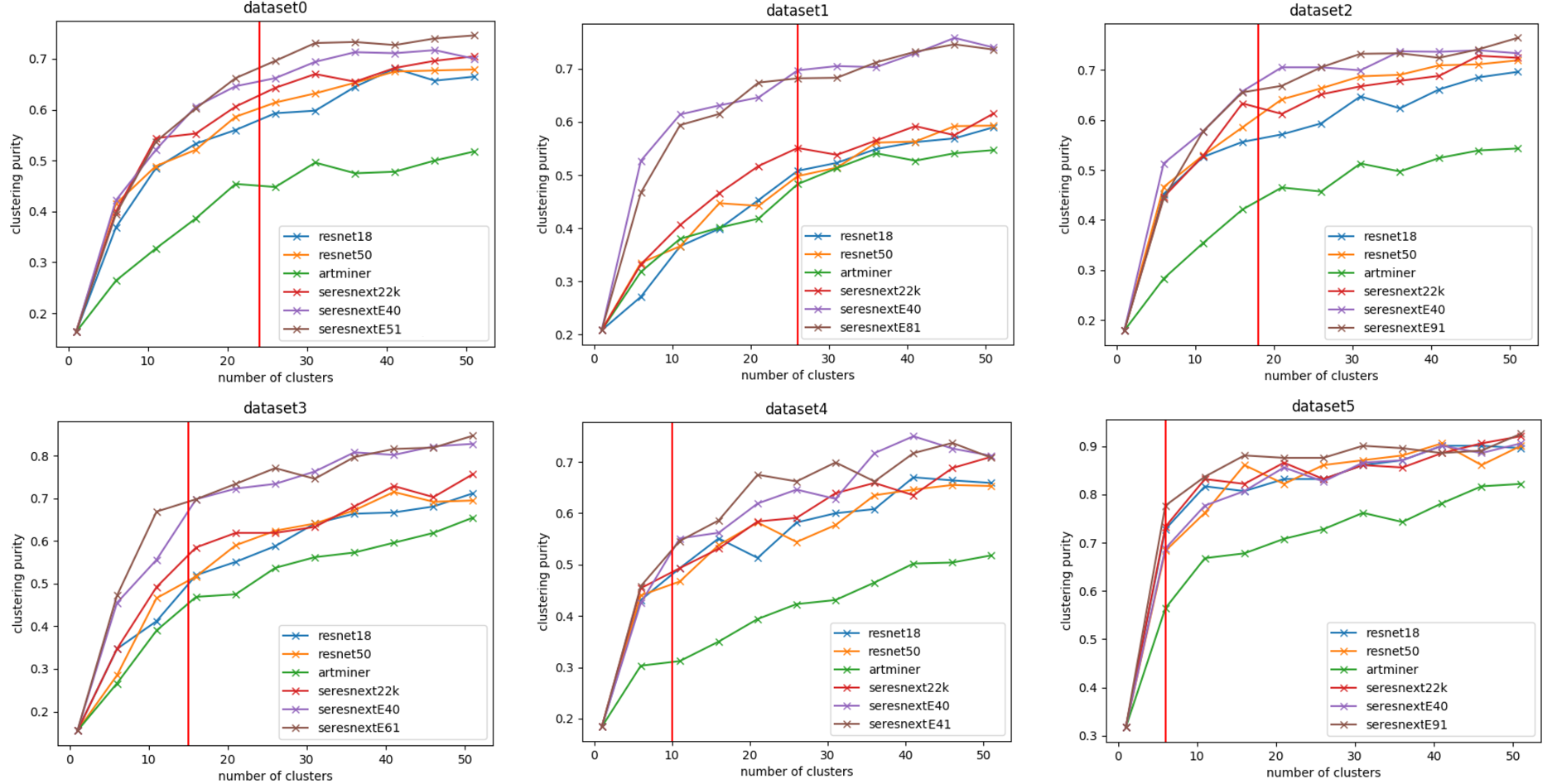}
\caption{Comparison of various network architectures for cluster purity on six different character subsets or as named in the figure - datasets. These character subsets were selected at random from the \textsc{CAST Test} data-set. Each graph shows the clustering purity measure as a function of the number of clusters for the different embedding network configurations. The red vertical line indicates the true number of characters (i.e. the number of clusters needed if they were all pure). Our network with a SEResNeXt architecture yields the best results across all data-sets, and this network seems to converge after 40 epochs.}
\label{purity_per_dataset}
\end{figure*}

We have analyzed the different alternatives for semantic embedding networks by evaluating their gain to cluster purity. As stated in the paper, the main goal of clustering is to reduce the load of user interaction in naming characters as much as possible. Hence, we prefer cluster purity over other measures. In Figure \ref{purity_per_dataset} we compare the performance of SEResNeXt, ResNet18, and ArtMiner for cluster purity. We show that our network based on a SEResNeXt backbone with 40 epochs of refinement converge to the top performance. This analysis does not include the \textsc{CAST} refinement.

\begin{figure*}
    \centering
    \includegraphics[width = 17cm]{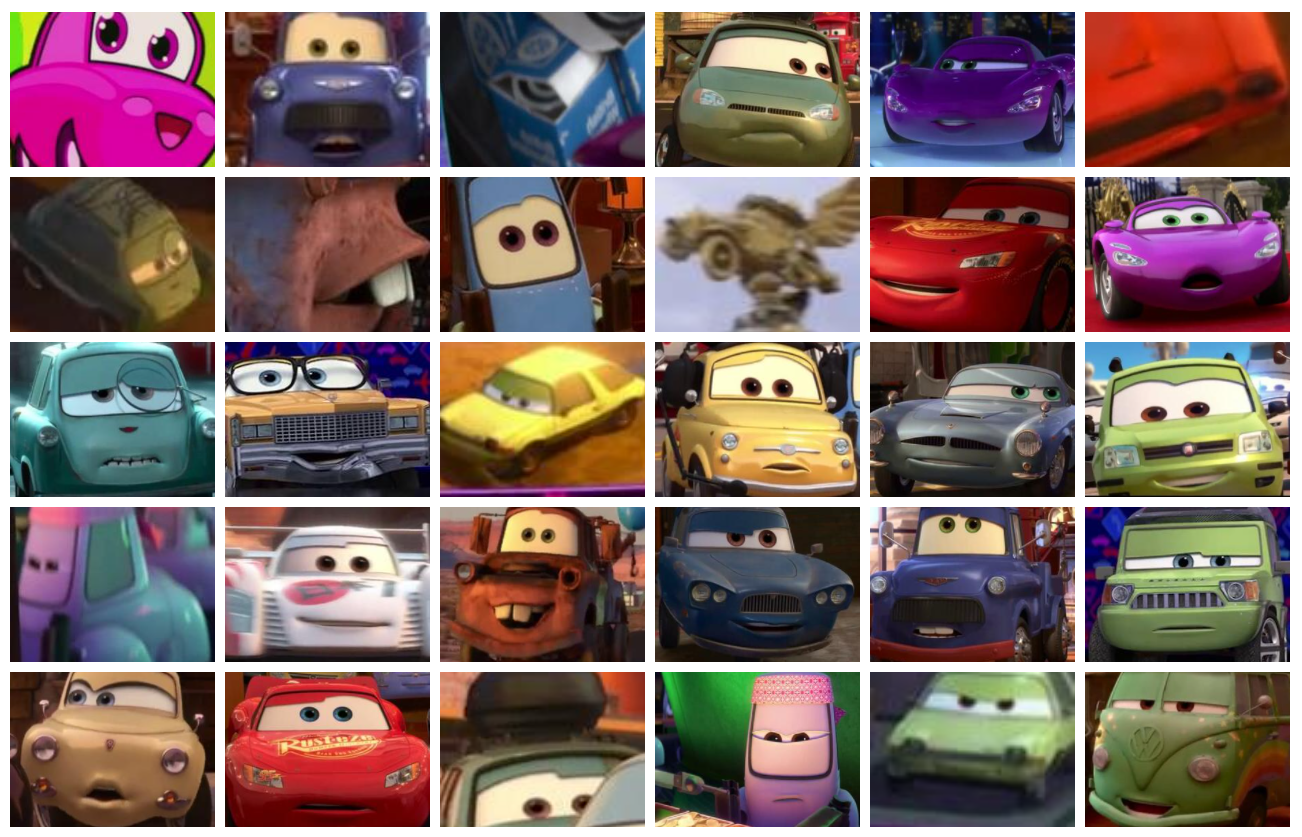}
\caption{Our method's dictionary output on the \textsc{SAIL AMCD} Evaluation video - `Cars 2'.}
\label{fig:sailCarsDict}
\end{figure*}

\begin{figure*}
    \centering
    \includegraphics[width = 17cm]{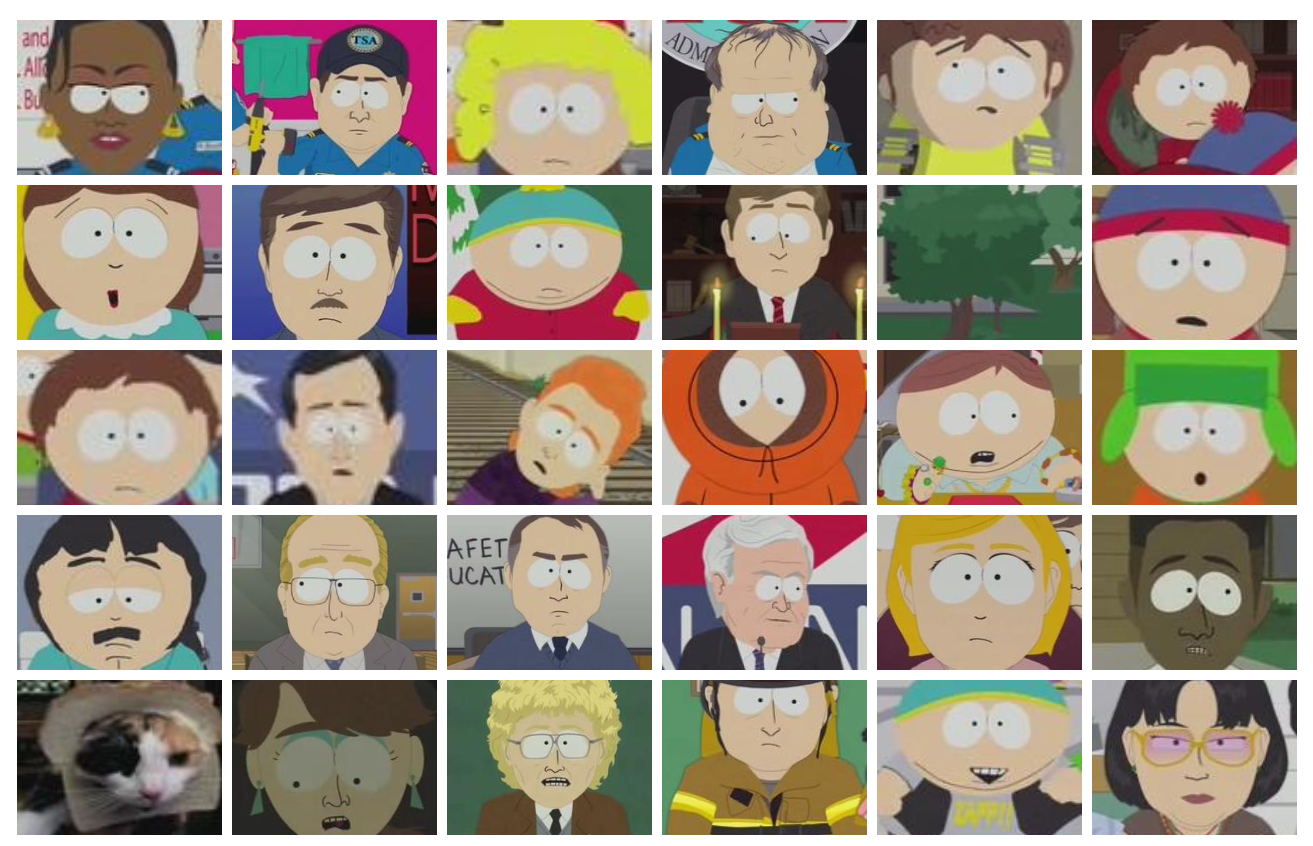}
\caption{Our method's dictionary extraction over the \textsc{CAST Evaluation} video - `Southpark'.}
\label{fig:sailCarsDict}
\end{figure*}

\section{Unsupervised discovery of character dictionaries}
To complete the visual picture of dictionary discovery by our method, on both \textsc{CAST} and \textsc{SAIL} datasets, the `Southpark' dictionary could be seen in Figure~\ref{fig:sailCarsDict}. Our method discovers 13 additional characters and considered relevant exemplars.
Another example for a dictionary discovery could be seen in Figure~\ref{fig:sailCarsDict}

\section{Character Identification}
\begin{figure*}
    \centering
    \includegraphics[width = 16cm]{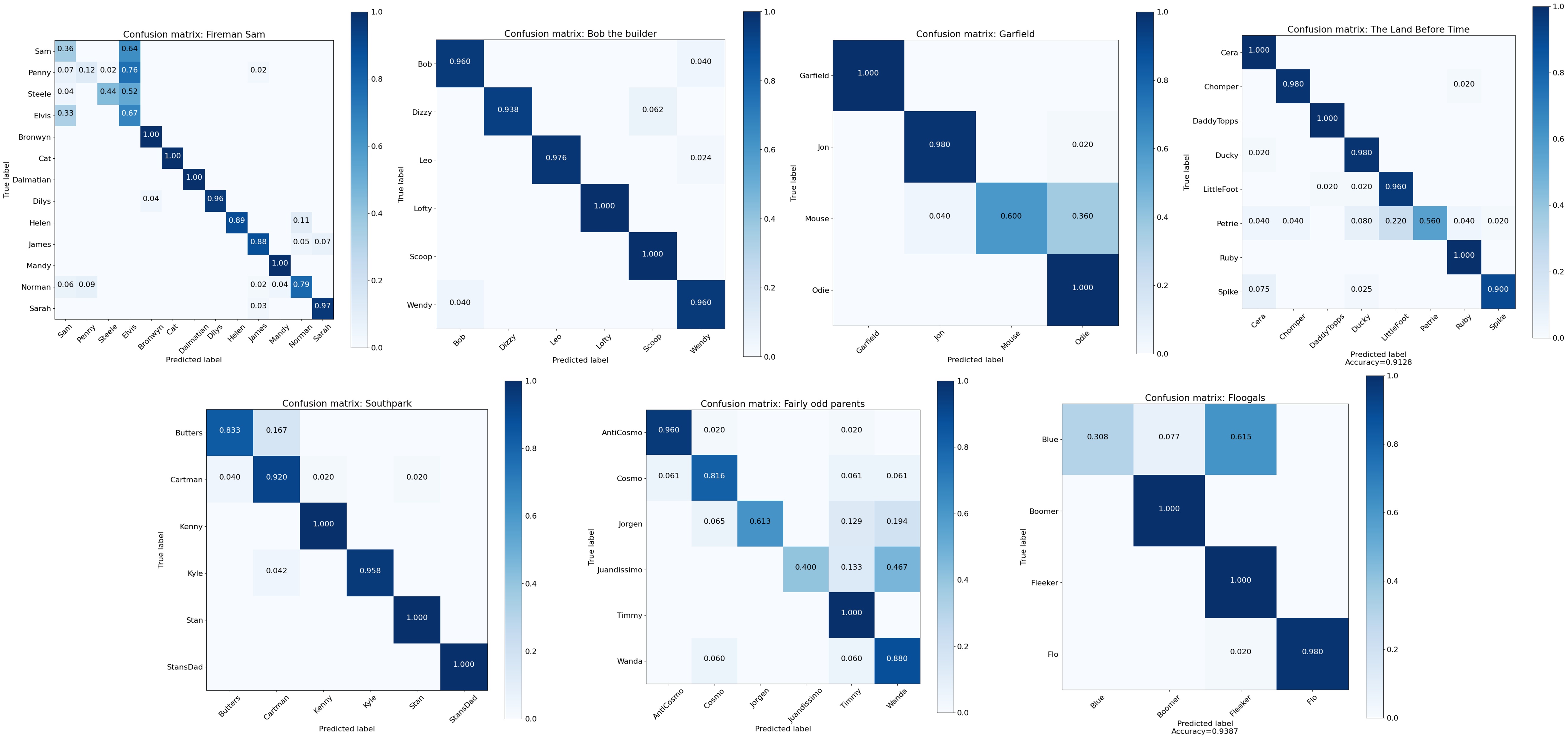}
\caption{The \textsc{Evaluation} dataset full Confusion Matrices}
\label{All_CMs_Evaluation}
\end{figure*}

We include all of the confusion matrices of the seven test-episodes' characters from \textsc{CAST Evaluation} data-set in Figure~\ref{All_CMs_Evaluation}. 

\begin{figure}
    \centering
    \includegraphics[width = 8.0cm]{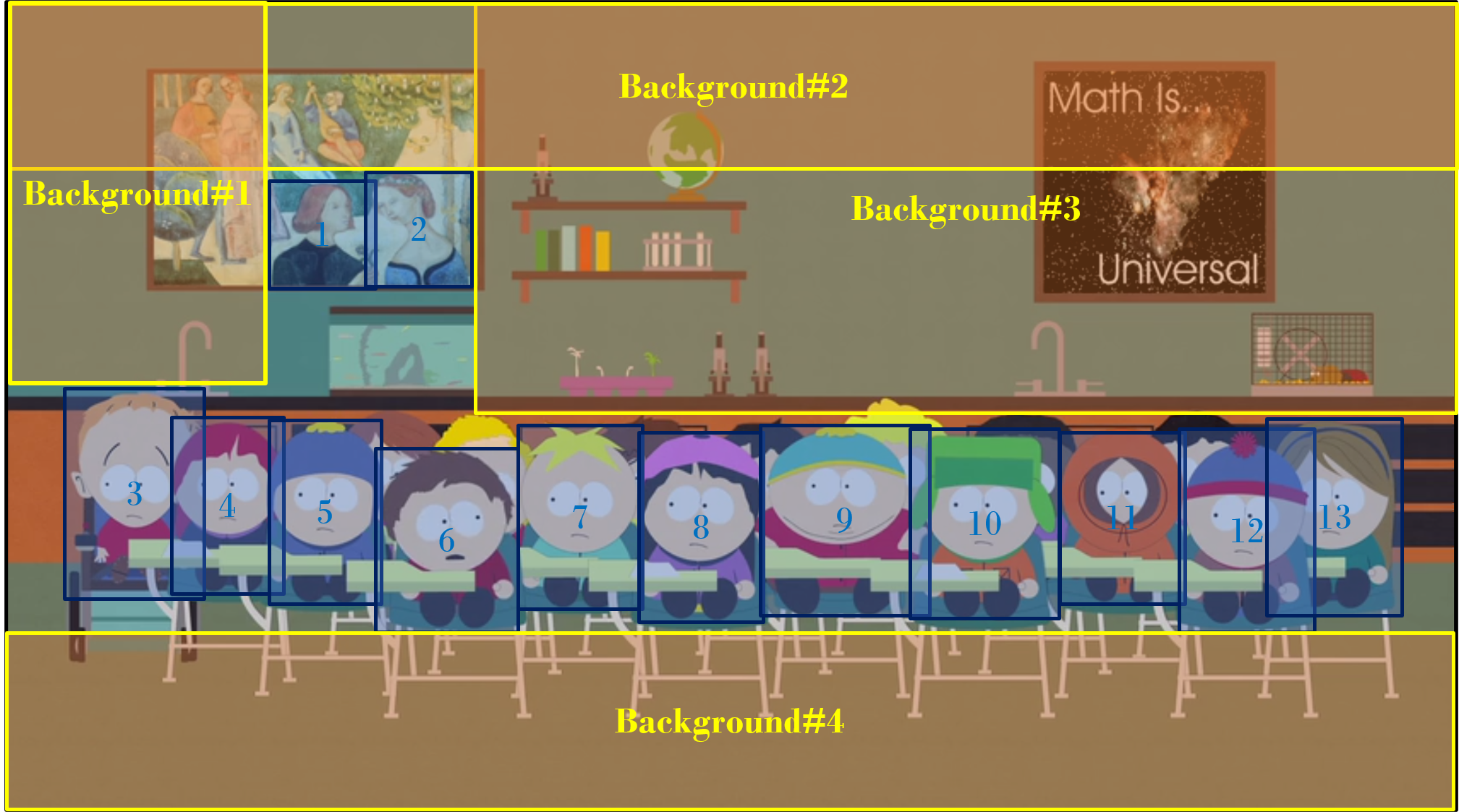}
\caption{Negative examples sampling from a video frame using our algorithm to find largest rectangles not including character proposals (`Southpark').}
\label{negative_sampling}
\end{figure}

\subsection{Negative examples sampling algorithm analysis}


 \begin{algorithm}
 \caption{LER($frame, boxes, min\_size$)}
 \begin{algorithmic}[1]
 \renewcommand{\algorithmicrequire}{\textbf{Input: Characters bboxes in a given frame}}
 \renewcommand{\algorithmicensure}{\textbf{Output: Large enough empty rectangles}}
 \REQUIRE character boxes
 \ENSURE  empty boxes
 \\ \textit{Initialisation} : boxes are consolidated and validated
  \IF {$\neg frame.is\_valid(min\_size)$}
  \RETURN $\{\}$
  \ENDIF

\IF{$|boxes| == 0$}
\RETURN $\{frame\}$
\ENDIF
  
  \STATE $\Theta \leftarrow \{\}$
  \STATE $center \leftarrow get\_centered\_bbox(boxes)$
  \STATE $subframes \leftarrow split\_by\_box(frame, center)$
  
  \FOR {$\sigma \in subframes$}
  \STATE { $\Theta \leftarrow \Theta \cup LER(\sigma, boxes, min\_size)$}
  
  \ENDFOR
 \RETURN $\Theta$ 
 \label{algo:largest_empty_rect}
 \end{algorithmic} 
 \end{algorithm}

The suggested method $LER()$ in Algorithm~\ref{algo:largest_empty_rect} is claimed to have a time complexity of $O(n^2)$. The proof is straight forward using the Master method. As illustrated in the pseudo code, the recursion has four splits into the top, bottom, left, and right parts of the frame where each is bounded by approximately one half of the frame size. On each call, the recursion handles the remaining boxes while filtering those who are outside the sub-frame area and cropping those that intersect with it. This operation is done in $O(n)$ time. 
Thus, the recursive work is bounded by the following function: $T(n)=4T(n/2)+O(n)$ which fits the first case of the Master method: $n^{\log_b a}$ where $a=4;b=2;d=1;$ which puts us
in a complexity of $O(n^2)$. QED.
See a visualization of the resulting output on Figure~\ref{negative_sampling}.

\subsection{Error analysis}
\begin{figure}
    \centering
    \includegraphics[width = 0.9\linewidth]{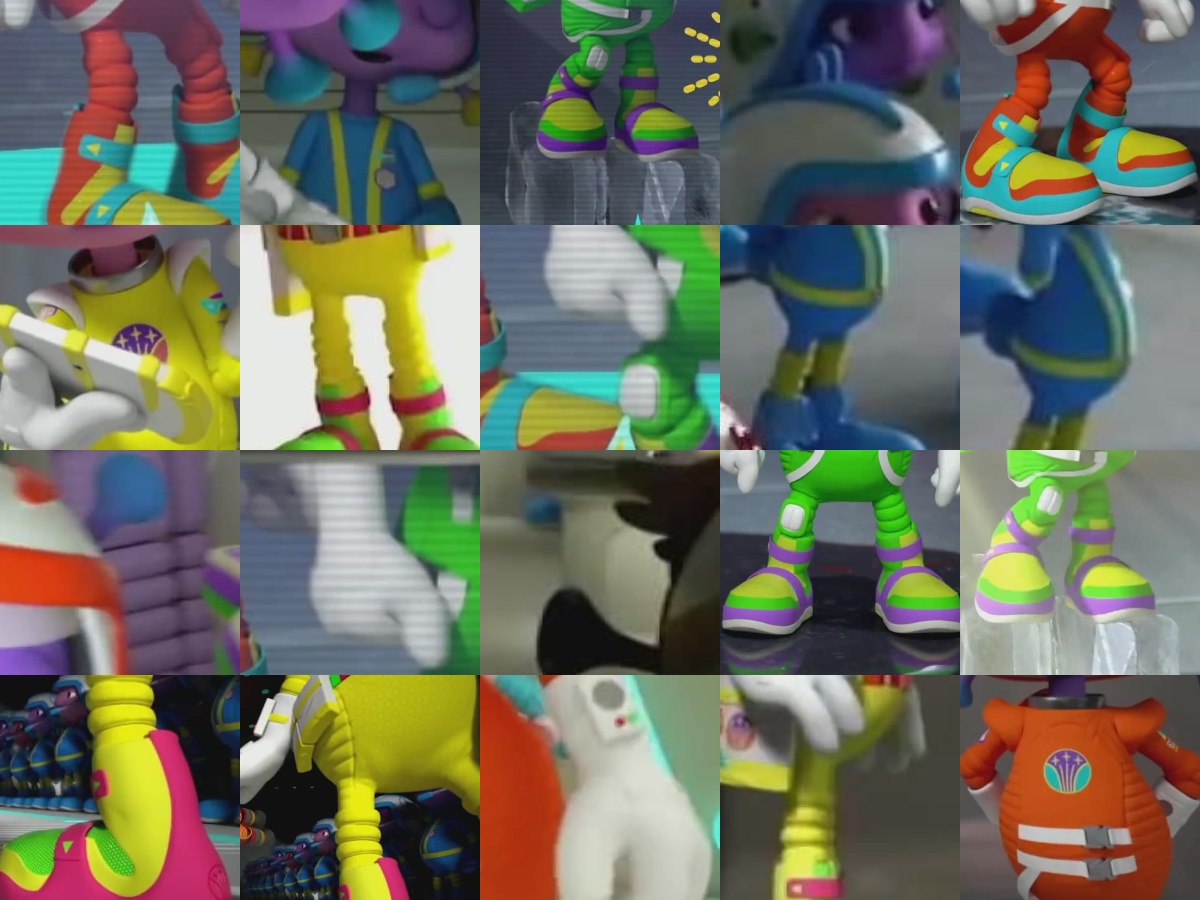}
\caption{False-negative examples include many character sub-parts (Floogals).}
\label{fig:Floogals_FalseNegatives}
\end{figure}

We further tested our detector by annotating \emph{all} proposal bounding boxes in the test-episodes of the series as ground truth. In this case, many proposals do not contain one of the characters and are marked as `unknown'. Error analysis on the false-negative cases revealed that many of them contained character parts such as hands, legs etc. These proposals were identified and annotated as characters by the human annotator, but are still a challenge for automatic classifiers (see Figure~\ref{fig:Floogals_FalseNegatives}). This may be an avenue of future research on identifying partial and occluded animated characters and the human factors in a setup of consistent data annotation in scale.

\end{document}